\pdfoutput=1

\documentclass[sigconf]{acmart}

\usepackage{algorithm}
\usepackage{algorithmicx}
\usepackage{graphicx}
\usepackage{subcaption}

\usepackage{multirow}

\usepackage{balance}

\def\BibTeX{{\rm B\kern-.05em{\sc i\kern-.025em b}\kern-.08emT\kern-.1667em\lower.7ex\hbox{E}\kern-.125emX}}
    
\copyrightyear{2019} 
\acmYear{2019} 
\setcopyright{acmcopyright}
\acmConference[CIKM '19]{The 28th ACM International Conference on Information and Knowledge Management}{November 3--7, 2019}{Beijing, China}
\acmBooktitle{The 28th ACM International Conference on Information and Knowledge Management (CIKM '19), November 3--7, 2019, Beijing, China}
\acmPrice{15.00}
\acmDOI{10.1145/3357384.3357956}
\acmISBN{978-1-4503-6976-3/19/11}

\begin{document}

\fancyhead{}

\title{Large Scale Landmark Recognition via Deep Metric Learning}

\author{Andrei Boiarov}
\affiliation{%
  \institution{Mail.ru Group}
  \streetaddress{Leningradsky prospect 39, bld. 79}
  \city{Moscow}
  \country{Russia}}
\email{a.boiarov@corp.mail.ru}

\author{Eduard Tyantov}
\affiliation{%
  \institution{Mail.ru Group}
  \streetaddress{Leningradsky prospect 39, bld. 79}
  \city{Moscow}
  \country{Russia}}
\email{tyantov@corp.mail.ru}

%
\renewcommand{\shortauthors}{Boiarov and Tyantov}

%
\begin{abstract}
This paper presents a novel approach for landmark recognition in images that we've successfully deployed at Mail.ru. This method enables us to recognize famous places, buildings, monuments, and other landmarks in user photos. The main challenge lies in the fact that it's very complicated to give a precise definition of what is and what is not a landmark. Some buildings, statues and natural objects are landmarks; others are not. There's also no database with a fairly large number of landmarks to train a recognition model. A key feature of using landmark recognition in a production environment is that the number of photos containing landmarks is extremely small. This is why the model should have a very low false positive rate as well as high recognition accuracy. 

We propose a metric learning-based approach that successfully deals with existing challenges and efficiently handles a large number of landmarks. Our method uses a deep neural network and requires a single pass inference that makes it fast to use in production. We also describe an algorithm for cleaning landmarks database which is essential for training a metric learning model. We provide an in-depth description of basic components of our method like neural network architecture, the learning strategy, and the features of our metric learning approach. We show the results of proposed solutions in tests that emulate the distribution of photos with and without landmarks from a user collection. We compare our method with others during these tests. The described system has been deployed as a part of a photo recognition solution at Cloud Mail.ru, which is the photo sharing and storage service at Mail.ru Group.
\end{abstract}

%
%


%

%

%

\maketitle

\section{Introduction}

\begin{figure}
  \centering
  \includegraphics[width=\linewidth]{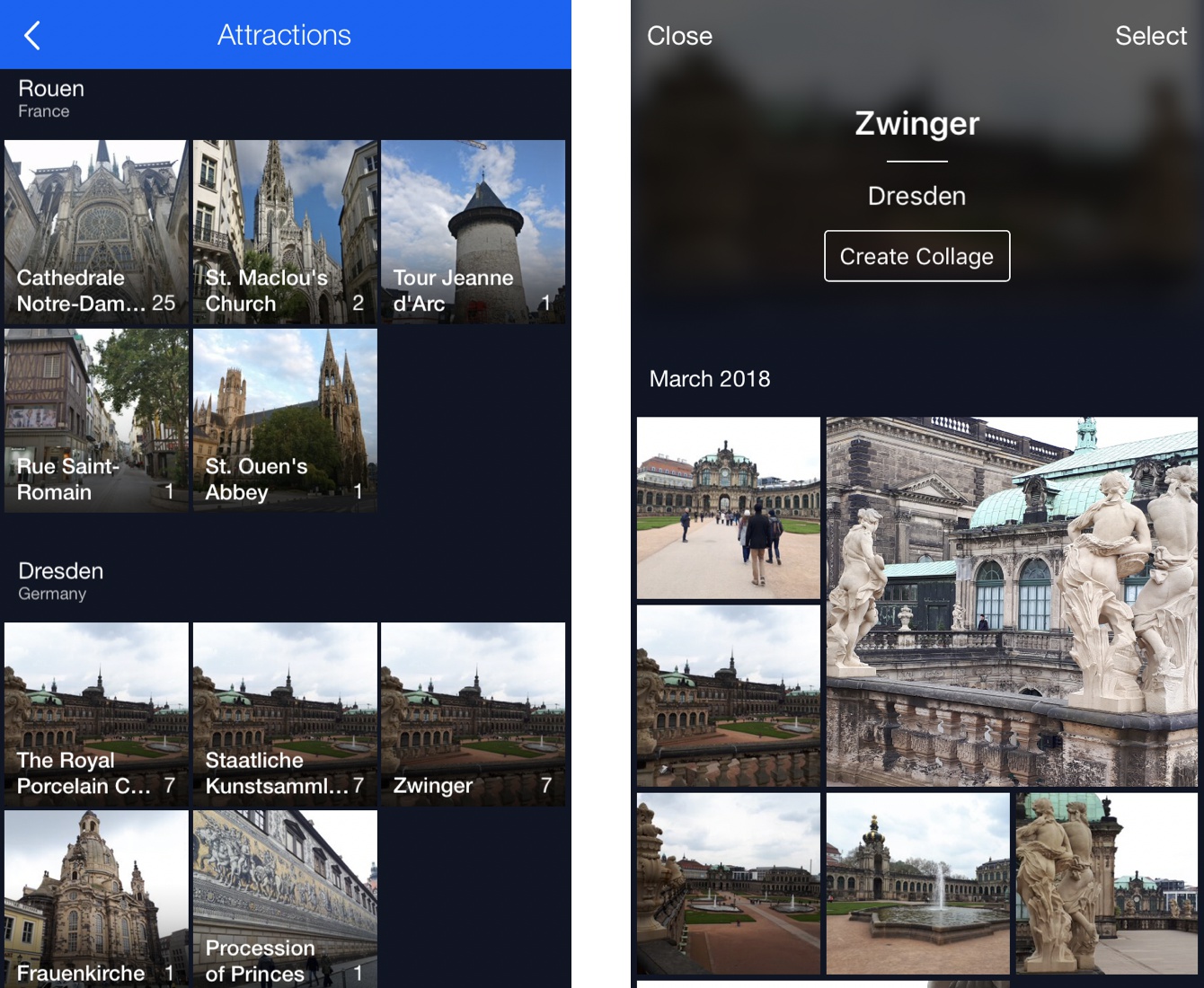}
  \caption{Example of user's photo landmarks (attractions) recognition in Cloud Mail.ru application. (Left:) Photos with landmarks grouped into albums by city (Rouen and Drezden). (Right:) Album containing all photos of the specific landmark (Zwinger).}
  \Description{Landmark recognition application example.}
  \label{fig:intro}
\end{figure}

By increasing amount of images in web and mobile applications, the issue of fast and accurate image search develops. A similar situation often arises in many applied areas such as e-commerce~\cite{yang2017visual, zhang2018visual} and search engine \cite{hu2018web}. These systems use various image retrieval techniques that allow extracted information to be used in searches.

Landmark recognition is an image retrieval task and has its own specific challenges. The first and most critical of them is that it is impossible to provide a concrete concept of a landmark. Landmarks do not have a common structure and can be just about anything: buildings, monuments, cathedrals, parks, waterfalls, castles, etc. What's more, we will find that some buildings are landmarks while others are not.

\begin{figure*}
  \centering
  \includegraphics[width=\textwidth]{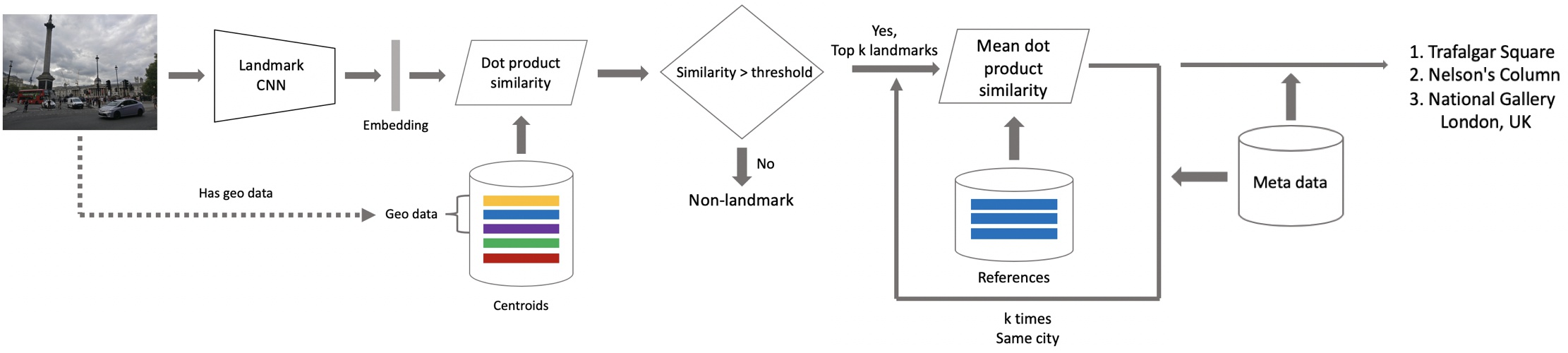}
  \caption{Overview of presented landmark recognition system. A query image is processed by the landmark CNN (Sections~\ref{cnn}, \ref{loss_section}, \ref{learning}), then resulted embedding vector is compared with precomputed centroids and filtered by class references and meta-data (Section~\ref{inference}).}
  \Description{System overview.}
  \label{fig:all_system}
\end{figure*}

The second problem is that a landmark can have different shooting angles and these angles can often be very different from each other. An extreme case is inside and outside the building. A good recognition system, however, should be able to identify a landmark from different shooting angles.

The third difficulty arises from the first, i.e. the fact that an overwhelming number of objects in the world are not landmarks. This leads to the requirement that a landmark recognition system must distinguish a large number of landmark classes among themselves with high accuracy. These classes are very skewed in relation to one large non-landmark class. Additionally, almost anything can be a non-landmark. This is why the system should have a very low false positive rate with high recognition accuracy.

We have developed and launched our landmark recognition system as part of our photo recognition solution at Cloud Mail.ru, the photo sharing and storage service at Mail.ru Group. We did this because it is important for users to be able to find landmarks among their photos. These photos are often special to users and they would like to be able to find them quickly. Our solution at Cloud Mail.ru allows photos with landmarks to be collected, sorted into a special album, and searched. An example of the landmarks album interface is shown in Figure~\ref{fig:intro}.

Training the landmark recognition model is challenging. There is no clean and sufficiently large open database that is suitable for what we need. The landmark recognition system must be sufficiently fast and scalable to be used in production. Additionally, user photos sometimes contain geo information about the location of where photos were taken. An important feature of a landmark recognition system is that it must show high quality without any additional information like geo information.

In this paper we present a landmark recognition system that successfully deals with these challenges and has been deployed as a part of the photo recognition solution at Cloud Mail.ru. Main contributions:
\begin{itemize}
    \item Inference using centroids that have resulted from hierarchical clustering; 
    \item Curriculum learning technique;
    \item Modified Center Loss~\cite{wen2016discriminative} for non-landmarks combined with a deep convolutional neural network pre-trained on a scenes database.
\end{itemize}
    
Main results:
\begin{itemize}
    \item Obtaining similar metrics to the state of the art model in our test;
    \item Inference time of our model is 15x faster;
    \item Efficient architecture enables our system to serve in an online fashion.
\end{itemize}

The rest of the paper is organized as follows: Section~\ref{sec:related} reviews recent
literature on landmark recognition algorithms. This is followed by Section~\ref{sec:approach}, in which deep convolutional neural network (CNN) architecture, loss function, training methodology, and aspects of inference are described. In Section~\ref{database}, we present the landmark database used to train our model as well as the method for cleaning this database. Next, in Section~\ref{experiments}, we show the results of quantitative experiments from our tests and analysis of the deployment effect to prove the effectiveness of our mode. Finally, results of comparison of proposed in this paper approach with the state-of-the-art model is presented in Section~\ref{sec:comparison}, followed by conclusion in Section~\ref{sec:conclusion}.

\section{Related work} \label{sec:related}

Landmark recognition can be considered as one of the tasks of image retrieval. The bulk of the literature on these topics is focused on the image retrieval task, while less attention is paid directly to the landmark recognition. The comprehensive survey of recent methods for these problems is presented by Radenovi\'{c}~{\it et al.}~\cite{radenovic2018revisiting}. We give here only the most influential. 

Over the past two decades, there has been significant progress in the field of image retrieval, and the main approaches can be divided into two groups. The first one is classical retrieval approaches using local features, such as: methods based on local features descriptors organized in bag-of-words~\cite{sivic2003video}, spatial verification~\cite{philbin2007object}, Hamming embedding~\cite{jegou2008hamming}, query expansion~\cite{chum2007total}. These approaches dominated in image retrieval until the rise of deep convolutional neural networks (CNN)~\cite{krizhevsky2012imagenet}, which are used to produce global descriptors for an input image. According to~\cite{radenovic2018revisiting}, the pipeline of a typical local-features based method includes local features detection, descriptor extraction, quantization using a visual codebook and aggregation into embeddings.  

The main representatives of local features-based methods include Vector of Locally Aggregated Descriptors~\cite{jegou2010aggregating} which uses first-order statistics of the local descriptors. Another method from this group is Selective Match Kernel~\cite{tolias2016image} --- an extension of the Hamming embedding approach.

Recently, with the advent of deep convolutional neural networks, the most efficient image retrieval approaches are based on training task-specific CNNs. Deep networks have proven to be extremely powerful for semantic feature representation, which allows us to efficiently use them for landmark recognition. As described in~\cite{radenovic2018revisiting}, the main CNN based approaches differ in various types of pooling layers, multi-scaling of an input image, usage of pre-trained models and pre-processing.

Metric learning aims at finding appropriate similarity measure between image pairs that preserve desired distance structure. A good similarity can boost the performance of image search, especially when the number of classes is large or unknown~\cite{bhatia2015sparse}. Gordo~{\it et al.}~\cite{gordo2017end} use regional max-pooling and fine-tuning by Triplet loss~\cite{schroff2015facenet}. Radenovi\'{c}~{\it et al.}~\cite{radenovic2018revisiting} employ Contrastive loss~\cite{hadsell2006dimensionality} and generalized mean-pooling, this approach shows good results, but brings additional memory and complexity cost~\cite{radenovic2018revisiting}.

DELF (DEep Local Features) by Noh~{\it et al.}~\cite{noh2017largescale} demonstrates promising results. This method combines classical local features approach with deep learning. DELF uses features from one of the layers of the deep CNN and train an attention module on top of it. This allows us to extract local features from input images and then perform geometric verification with RANSAC~\cite{fischler1981random}. As follows from the experiments described in~\cite{radenovic2018revisiting}, DELF and its extension~\cite{teichmann2018detect} are the state-of-the-art approaches. 

Our goal in this paper is to describe the method that uses strengths of deep convolutional neural networks for accurate and fast large scale landmark recognition.

\section{Approach} \label{sec:approach}
In this section we describe in detail our landmark recognition system, outlined in Figure~\ref{fig:all_system}. Like many state-of-the-art image retrieval methods~\cite{noh2017largescale, gordo2017end, radenovic2018fine} ours is based on deep convolutional neural network (CNN). For inference we use prototypical based approach where all classes are characterized by corresponding centroids. This type of metric learning combines speed and accuracy for the large scale recognition task. In the following we describe architecture, loss function, training methodology and aspects of inference.

\begin{table}
  \caption{Scenes database validation errors}
  \label{tab:scenes}
  \begin{tabular}{lcc}
    \toprule
    Model&Top 1 (\%)&Top 5 (\%)\\
    \midrule
    ResNet-50 \cite{he2016deep} & 46.1 & 15.7 \\
    ResNet-200 \cite{he2016deep} & 42.6 & 12.9\\
    SE-ResNext-101 \cite{hu2018squeeze} & 42 & 12.1 \\
    {\bf WRN-50-2 \cite{zagoruyko2016wide}} & {\bf 41.8} & {\bf 11.8}\\
  \bottomrule
\end{tabular}
\end{table}

\subsection{CNN architecture} \label{cnn}

Recent success in computer vision and image retrieval are closely related to convolutional neural networks~\cite{krizhevsky2012imagenet}. Our landmark CNN consists of three parts: the main network, embedding layers and the classification layer.

First of all, we chose the optimal network architecture.
In order to get a CNN suitable for training landmark recognition model, we used fine-tuning: a very efficient technique in various computer vision applications. 
Landmark and scene recognition tasks are closely related, since almost all landmarks can be divided into groups by scenes. So we chose to pre-train the network on our scenes database, which was previously used for training scene recognition as part of tagging solution at Cloud Mail.ru. The database is inspired by the Places database~\cite{zhou2018places} and contains about $1.5$ million images from 314 scene categories collected from our internal data. Scenes data provides the CNN with necessary signal for learning general concepts about these groups. Comparison of different Residual Networks~\cite{he2016deep} can be found in Table~\ref{tab:scenes}. Based on the results, we chose Wide Residual Network (WRN-50-2)~\cite{zagoruyko2016wide}. This network architecture showed good trade-off between quality and inference time. 

Secondly, we add the embedding layer to the CNN trunk. We cut off last fully-connected layers after average pooling of the network and add fully-connected layer and batch normalization~\cite{ioffe2015batch} instead. Embedding layer size is an important parameter. After a series of experiments on validation data (see Table~\ref{tab:offline_hyperparams} for results on our offline test (Section~\ref{experiments})), we chose $512$ as optimal.
The third component of the landmarks CNN is a classification fully-connected layer. Full architecture of the landmark CNN is presented in Figure~\ref{fig:cnn}.

\begin{figure}
  \centering
  \includegraphics[width=\linewidth]{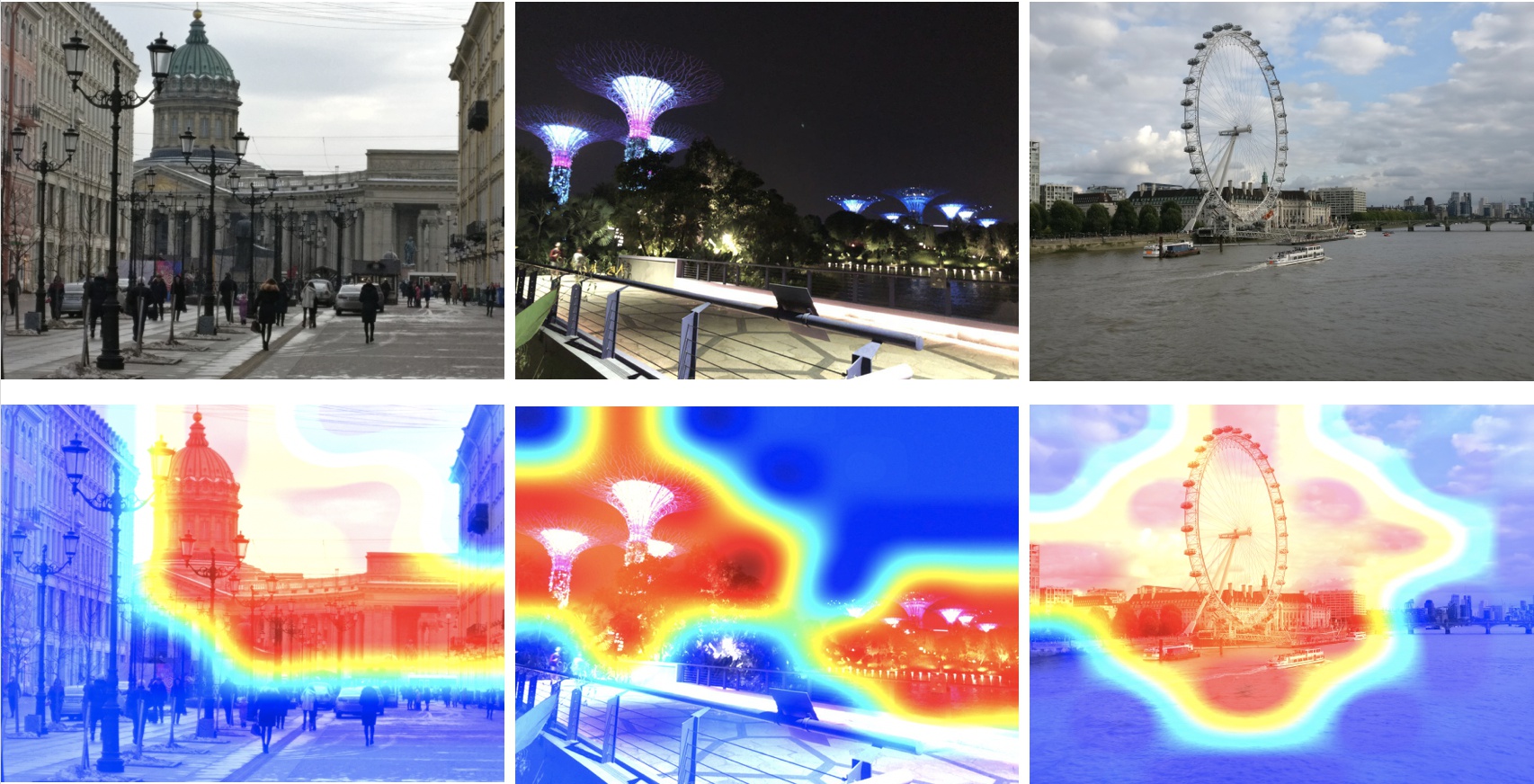}
  \caption{Examples of class activation maps generated by largest values of the embedding layer. All images are correctly recognized. (Left column): Kazan Cathedral in St. Petersburg; (Middle column): Supertree Grove is Singapore; (Right column): London Eye.}
  \Description{CAM examples.}
  \label{fig:cam}
\end{figure}

To check what the CNN has learned we use class activation maps (CAM)~\cite{zhou2016learning}, obtained by the largest value of activations from the embedding layer. The examples of CAMs for the CNN trained on our landmark database are presented in Figure~\ref{fig:cam}. As can be seen, the CNN highlights parts of the images with the correct landmark.

\subsection{Loss function} \label{loss_section}

As mentioned earlier, our landmark recognition pipeline during the inference (see section \ref{inference}) uses centroids that describe each landmark class. In order to maximize the effectiveness of this approach, we had to train the network to have the members of each class to be as close as possible to some element we call the ``center'' (or ``centroid''). We experimented with couple suitable loss functions: Contrastive loss~\cite{hadsell2006dimensionality}, Arcface~\cite{deng2018arcface} and Center loss~\cite{wen2016discriminative}. The best results on our tests were achieved using Center loss, which simultaneously trains a center for embbedings of each class and penalizes distances between image embbedings and their corresponding class centers. A significant advantage of Center loss is that it's just simple addition to Softmax loss and, unlike Contrastive loss or Triplet loss~\cite{schroff2015facenet}, there is no need to sample negative pairs, which can be quite challenging.

An important feature of landmark recognition is that query photos may not contain any landmarks at all, so our loss function must take this into account. Thus, we consider an additional class of {\it non-landmark} which includes literally everything that is not a landmark: persons, animals, buildings, natural objects, cars etc. Therefore, the number of elements of non-landmark class is much greater than the total number of elements in all landmark classes. According to our statistics of user photos, the number of photos with landmarks is about $1-3\%$ of the total number of photos. In addition, unlike landmarks, non-landmark class does not have structure (since it includes all sorts of objects), so it makes no sense to enforce Center loss on this class. In order to address this issue, we modified Center loss to calculate centers only for landmark classes, but not for the non-landmark class. 

Let $n$ be the number of landmark classes, $m$ --- mini-batch size, $\mathbf{x}_i \in \mathbb{R}^d$ - the $i$th embedding and $\mathbf{y}_i$ --- the correct label, $\mathbf{y}_i \in \{1, \ldots, n, n+1 \}$, where $n+1$ is the label of a non-landmark class. Denote $W \in \mathbb{R}^{d \times n}$ as the weights of the classifier layer, $W_j$ as its $j$th column. Let $\mathbf{c}_{\mathbf{y}_i}$ be the $\mathbf{y}_i$th embeddings center from Center loss and $\lambda$ - the balancing parameter of Center loss. Then our loss function, which we minimize during the training process, is formulated as:

\begin{equation} \label{loss}
    \mathcal{L} = - \sum_{i=1}^m \log \frac{\exp^{W^\intercal_{\mathbf{y}_i} \mathbf{x}_i}}{\sum_{j=1}^{n+1} \exp^{W^\intercal_j \mathbf{x}_i}} + \frac{\lambda}{2} \sum_{\substack{i=1\\ \mathbf{y}_i < n+1}}^m \| \mathbf{x}_i - \mathbf{c}_{\mathbf{y}_i} \|_2^2.
\end{equation}

\begin{figure}
  \centering
  \includegraphics[width=\linewidth]{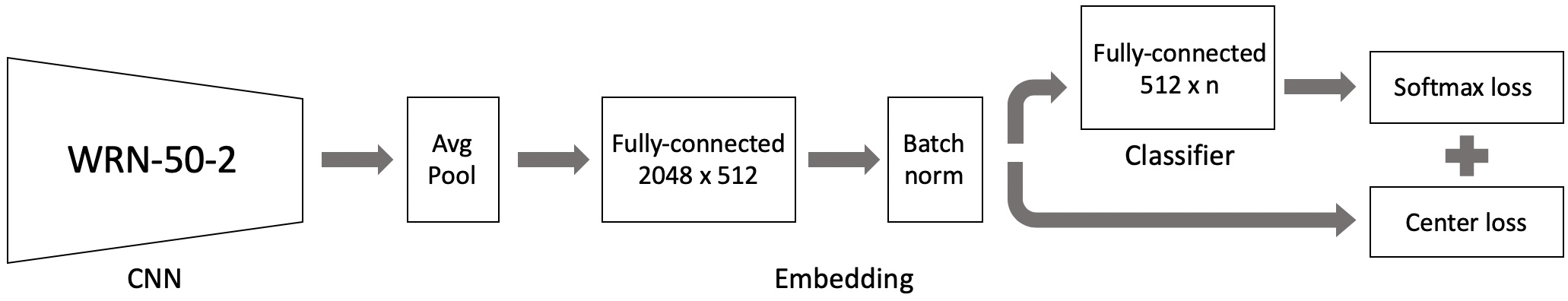}
  \caption{Architecture of the deep convolutional neural network for landmark recognition and losses.}
  \Description{CNN learning architecture.}
  \label{fig:cnn}
\end{figure}

Our landmark CNN with described loss function is shown in Figure~\ref{fig:cnn}. The advantage of using Center loss is presented in Table~\ref{tab:offline_no_geo}.

\subsection{Training} \label{learning}

In this section we describe the tricks that we use for training the model, presented in Section~\ref{cnn} with loss function given by Eq.~\ref{loss}.

\subsubsection{Learning parameters}
As optimization algorithm we use stochastic gradient decent with momentum~\cite{rumelhart1986learning}, during training the momentum value is $0.9$. Our CNN is trained with weight decay parameter $5e-3$ and Center loss parameter $\lambda=5e-5$. Each training image is resized to $256 \times 256$ pixels, and then data augmentation techniques such as random resized crop~\cite{schroff2015facenet} to $224 \times 224$, color jitter and random flip are applied. As described in Section~\ref{cnn}, our landmark CNN consists of three parts: WRN-50-2 pre-trained on the scenes database, embedding layers and a classifier. For each part we use different learning rates: $\alpha_1, \alpha_2, \alpha_3$ accordingly. The CNN was trained for $30$ epochs and learning rates decay at $0.1$ twice after $40\%$ and $80\%$ of the total number of learning epochs.

\subsubsection{Curriculum learning} \label{seq:curr_learnig}
\begin{figure}
  \centering
  \includegraphics[width=0.8\linewidth]{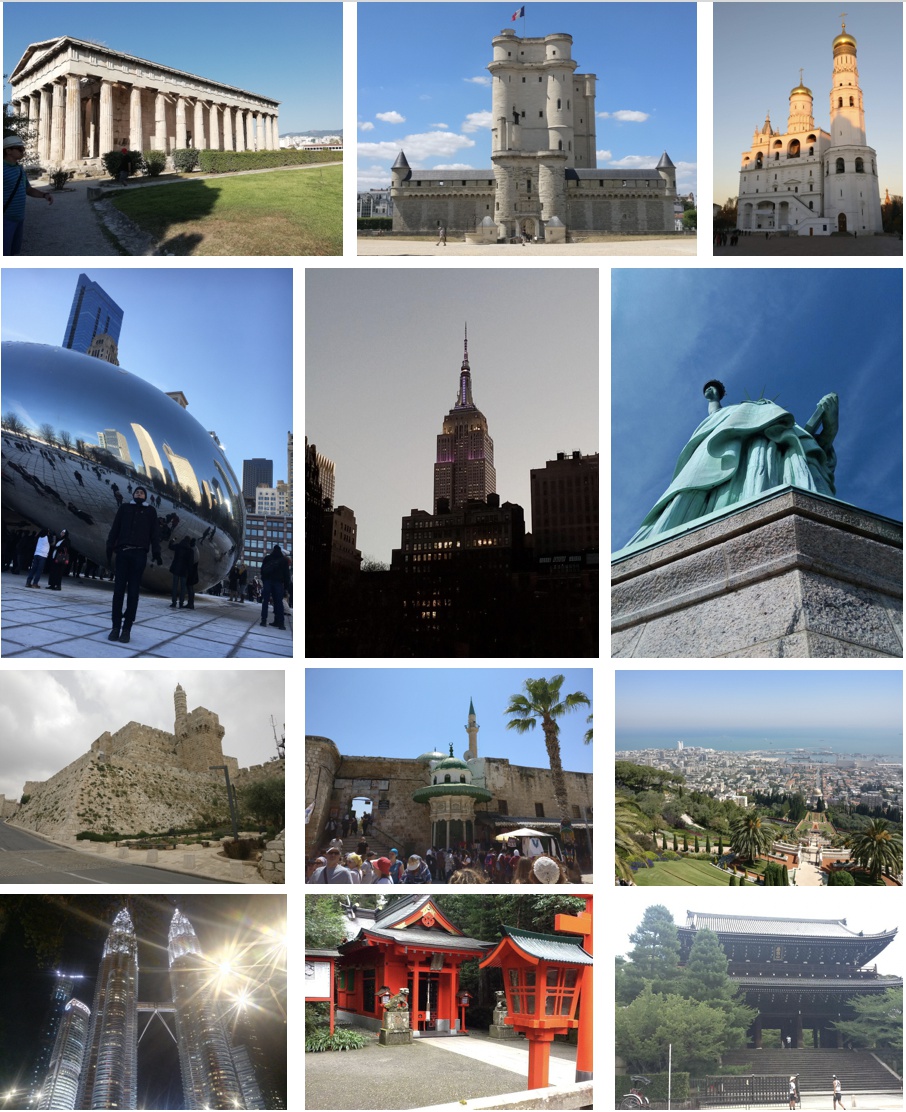}
  \caption{In rows: representatives of four parts of the landmark database for curriculum learning.}
  \Description{Database parts.}
  \label{fig:data_parts}
\end{figure}

In the beginning of our experiments we tried training CNN on all the landmark data, but this approach showed poor results (see Table~\ref{tab:offline_no_geo}). Significantly more successful was the strategy inspired by a curriculum learning ~\cite{bengio2009curriculum}, that made it possible to achieve high recognition accuracy. Under a curriculum learning approach, the landmark database was divided into parts according to the geographical affiliation of cities where landmarks are located:
\begin{enumerate}
    \item Europe (including all Russia);
    \item North America, Australia and Oceania;
    \item Middle East, North Africa;
    \item Far East.
\end{enumerate}
Typical representatives of these parts of the database are presented in Figure~\ref{fig:data_parts}. The main motivation for using this approach is that landmarks from one geographical region are more similar to each other than landmarks from different regions. Thus, by training at landmarks from one region, the CNN learns information, which allows it to distinguish well these similar landmarks. 

Curriculum learning procedure consists of several steps, on each a new fraction of the database is added to the training. Full details are described in Algorithm~1.

\begin{algorithm} \label{alg:curr}
\caption{Curriculum learning for landmark recognition}
\begin{algorithmic}[1]
\Statex \textbf{Input:} Four parts of the landmark database; Landmark CNN.
\Statex \textbf{Output:} Trained model.
\State Train the landmark CNN on part (1) of the database with learning rates: $\alpha_1=0.001$, $\alpha_2=0.01$, $\alpha_3=0.01$.
\State Take the trained CNN. Batch normalization from embedding layers, classifier and centers $\mathbf{c}_{\mathbf{y}_i}$ are initialized randomly.
\State Train it on data from combining parts (1) and (2) of the landmark database. Learning rates: $\alpha_1=0.0001$, $\alpha_2=0.0001$, $\alpha_3=0.01$.
\State Repeat steps 2 and 3 for data obtained from parts $(1) + (2) + (3)$.
\State Repeat steps 2 and 3 for data obtained from parts $(1) + (2) + (3) + (4)$.
\State Return trained landmark CNN.
\end{algorithmic}
\end{algorithm}

\subsubsection{Non-landmarks influence}
As was discussed in Section~\ref{loss_section} the number of images without landmarks is crucial for the entire training process of the landmark recognition system. During our experiments on validation data, we increased the fraction of non-landmarks in the training data (see Table~\ref{tab:offline_hyperparams} for results on our offline test (Section~\ref{experiments})). As a result, we use the number of the non-landmark class and the total number of elements from landmark classes in proportion $0.4$.   

\subsection{Inference} \label{inference}

As mentioned in previous sections, we train our landmark CNN in such manner that efficiently uses prototypical approach during inference. This approach has been chosen because it proved to be stable at large scale during our experiments. The scheme of this stage is presented in Figure~\ref{fig:all_system}. In sections~\ref{cnn}, \ref{loss_section}, \ref{learning} architecture, loss function and training tricks are described in details. In the following sections we discuss centroids calculation, geo information, references and usage of meta data.

\subsubsection{Centroids}

In order to decide whether a query image contains a landmark, it is necessary to compare the embedding vector obtained from the landmark CNN with some others vectors describing classes of landmarks. We call these vectors {\it centroids}, because they are calculated by averaging embeddings. 

The main question about centroids is what data to use for its calculation. At first, we used the whole training data for each landmark. This approach yielded satisfactory results (see Table~\ref{tab:offline_knn}), but we began to look for a better method that would allow us to take into account peculiarities of our data (that is, landmarks). First, in spite of data cleaning (described in more detail in Section~\ref{database}), which removed most of the redundant elements, some of them remained in the data, which may affect the centroid. For example, if the landmark we are interested in is a palace which located on a city square, then images of a similar building on the same square may be included in the data. 

Second, each landmark can have several typical shooting angles. Mixing them in one centroid leads to its lack of sensitivity. It's more efficient to calculate a separate centroid for each typical shooting angle. To achieve this goal, we apply hierarchical agglomerative clustering algorithm~\cite{mullner2011modern} and partition training data into several clusters for each landmark.

We consider clusters as valid, if they contain more than $50$ elements. This number and clustering parameters have been chosen as a result of the experiments on validation set, maximizing classification metrics (see Table~\ref{tab:offline_hyperparams} for results on our offline test (Section~\ref{experiments})). As a result, we chose complete linkage clustering with a threshold equals $10$. Centroids for the current landmark are calculated using these valid clusters. If all clusters for the landmark are not valid then only one centroid is calculated using largest cluster. Let $v$ be the number of valid clusters for a landmark $l$, if there is no valid clusters for $l$ then $v=1$. Denotes $\mathbf{C}_j, j \in 1, \ldots, v$ as valid clusters (or one maximum cluster if there is no valid clusters), then set of centroids for a landmark $l$ $\{\mu_{l_j}\}_{j=1}^v$ is calculated by:    

\begin{equation} \label{eq:centroid}
    \mu_{l_j} = \frac{1}{|\mathbf{C}_j|} \sum_{i \in \mathbf{C}_j} \mathbf{x}_i, \; j \in 1, \ldots, v.
\end{equation}

\begin{figure}
  \centering
  \includegraphics[width=0.8\linewidth]{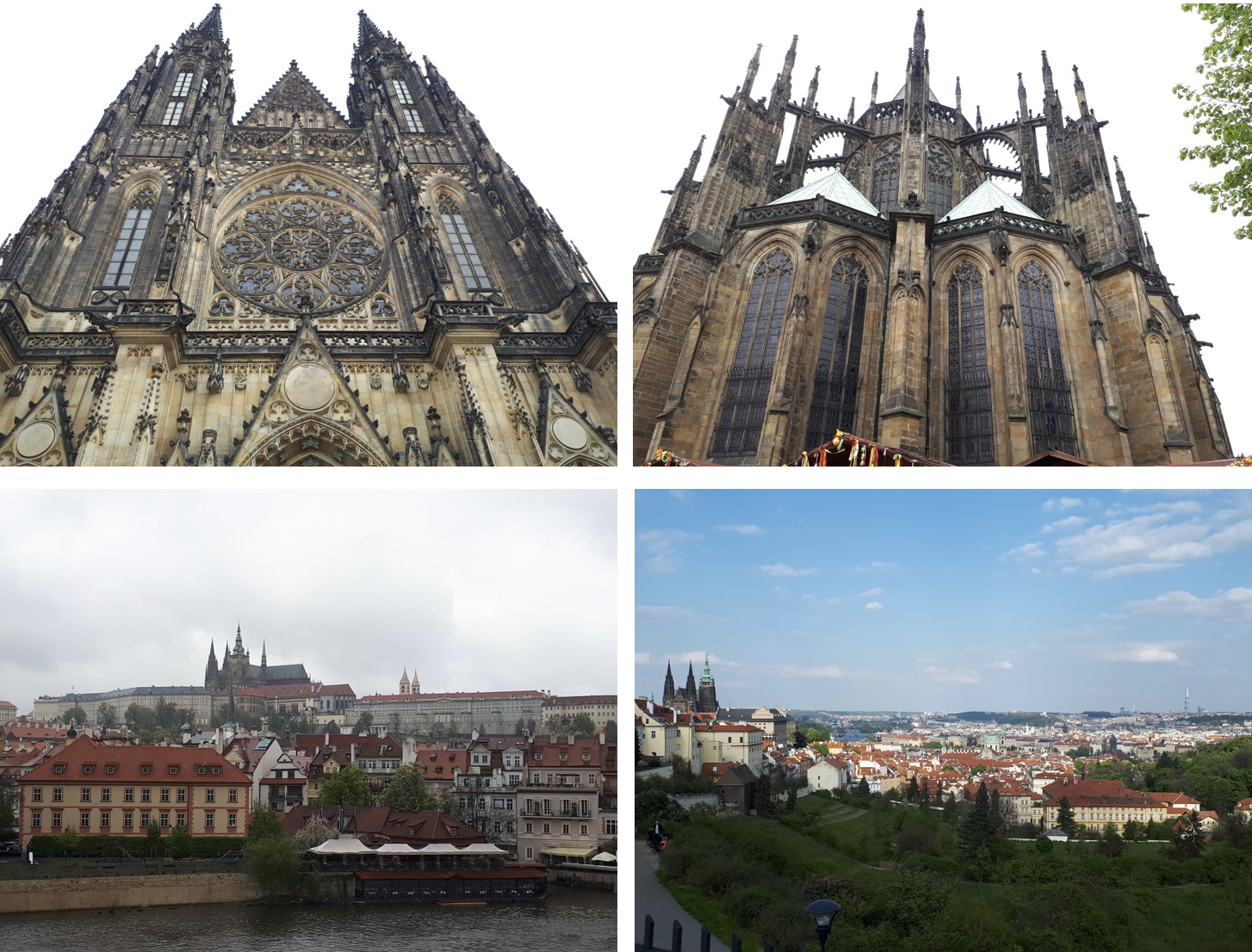}
  \caption{Church of St. Vitus in Prague Castle: closest images from user photos to the first centroid (top row) of the class and to the second centroid (bottom row).}
  \Description{Prague Castle several class centroids.}
  \label{fig:cent_1}
\end{figure}

\begin{figure}
  \centering
  \includegraphics[width=0.8\linewidth]{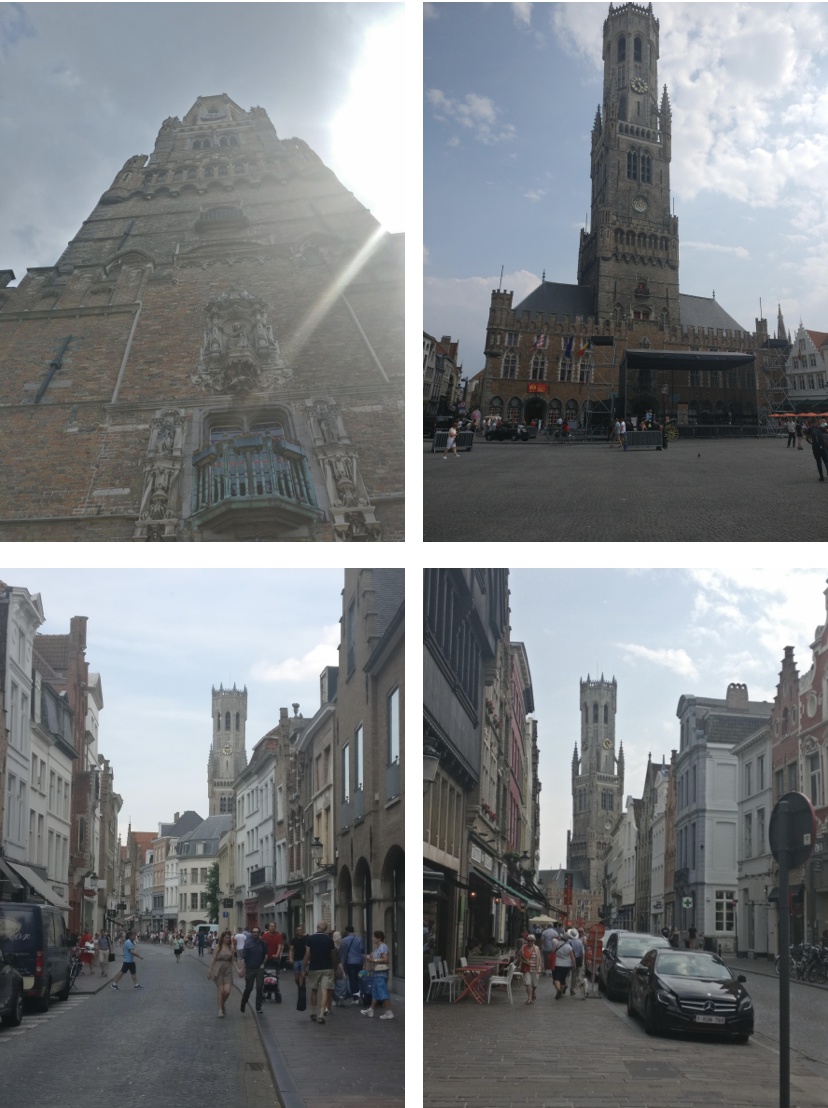}
  \caption{Belfort in Bruges: closest images from user photos to the first centroid (top row) of the class and to the second centroid (bottom row).}
  \Description{Belfort several class centroids.}
  \label{fig:cent_2}
\end{figure}

After calculating centroids for our landmark database (Section~\ref{database}) about $20 \%$ of the landmarks have more than one centroid per class (which corresponds to ``frequent'' landmarks with large number of photos). To illustrate that different centroids reflect different typical shooting angles of this landmark, we present Figures~\ref{fig:cent_1} and \ref{fig:cent_2}.

\subsubsection{Inference algorithm}

After centroids are calculated for all classes, we can make decision whether there are any landmarks on an image. To do this the query image is passed through the landmark CNN and resulting embedding vector is compared with all centroids by dot product similarity. Since there are many classes in our database, simple one by one comparison is inefficient. Hence, we employed approximate k-nearest neighbors (AKNN) search instead. For that we used very fast and popular Faiss~\cite{JDH17} library.

We experimented with k-nearest neighbors (kNN) approach for embeddings and centroids on validation data. Results of this experiments on our offline test (Section~\ref{experiments}) is presented in Table~\ref{tab:offline_knn}. In our model we use several centroids with $k=1$.

In order to distinguish landmark classes from non-landmark, a threshold value $\eta$ is introduced. If maximum similarity is less than $\eta$, then we decide that the image does not contain any landmarks. Else we take top $k$ similarities greater than $\eta$ and decide on the presence of the corresponding landmarks. This procedure is depicted in the middle part of Figure~\ref{fig:all_system}.

All landmarks found on the image must be in the same city. We use meta data from our landmark database (in more details Section~\ref{database}) to verify this.

If we can extract additional information from a query image, then we can improve the quality of recognition. Obviously, we can use geo data about the place where the photo was taken. It must be noted that users' photos do not always contain this data, therefore our main task was to develop a system that will produce high quality results in the absence of any additional information about the image. If a query image has geo information, then we can narrow the search range from all landmarks to landmarks lying in some surroundings from the shooting location. In this case query image embedding vector is compared only to centroids corresponding to landmarks from this surroundings, which allows us to reduce the threshold $\eta$, since very similar objects rarely lie next to each other. As surroundings, we use square $1 \times 1$ kilometers.

During inference, if an image doesn't have geo information, we perform additional verification step using landmark references (right part of Figure~\ref{fig:all_system}). After cleaning the database (Section~\ref{database}) there are $3-5$ manually checked images containing this landmark called references for each landmark. During the references verification step a query image embedding vector is compared to all references embedding vectors of the current landmark and, if mean dot product similarity is greater than another threshold $\omega$, then the landmark presence is considered confirmed. This approach allows us to slightly reduce false positive rate, which we found useful in practice. Since user photos don't always contain geo information, it's important to show how the described method works without any additional information. Therefore, in the following sections, test results are presented without usage of geo information.

A full description of the inference algorithm (with landmark references) is given in Algorithm 2.

\begin{algorithm}
\caption{Inference algorithm}
\begin{algorithmic}[1]
\Statex \textbf{Input:} Query image; Landmark CNN; Centroids matrix $\{\{\mu_{l_j}\}_{j=1}^v\}_{l=1}^n$ computed by Eq.~\ref{eq:centroid}; $N$ --- number of centroids; References; Metadata; $k$ is a maximum number of landmarks.
\Statex \textbf{Output:} List of found landmarks.
\State \textit{embedding} = CNN(image) --- pass query image through the landmark CNN and get embedding
\State \textit{scope} = $[1, \ldots, N]$
\State \textbf{if} (query image has geo information) \textbf{then}
\State \quad \textit{scope} = $[N_1, \ldots, N_2]$ --- indexes of the geographically nearest landmarks 
\State \textit{similarities} = kNN(\textit{embedding}, Centroids[\textit{scope}])
\State \textbf{if} (max(\textit{similarities}) < $\eta$) \textbf{then}
\State \quad \textbf{return} Non-landmark
\State \textit{landmark-indexes} = argsort(\textit{similarities})$[1, \ldots, k]$ --- indexes of $k$ top landmarks (by \textit{embedding})
\State \textit{city} = Metadata[$i$][city] --- city name for top landmark 
\State \textbf{for} ($i \in$ \textit{landmark-indexes}) \textbf{do}
\State \quad \textit{mean-similarity} = meanSimilarity(References[$i$], \textit{embedding})
\State \quad \textit{current-city} = Metadata[$i$][city]
\State \quad \textbf{if} (\textit{mean-similarity} > $\omega$ \textbf{and} \textit{current-city} == \textit{city}) \textbf{then}
\State \quad \quad \textbf{yield} Metadata[$i$][landmark-name], \textit{city}
\end{algorithmic}
\end{algorithm}

\section{Landmark database} \label{database}

In this section we describe the landmark database, on which our model was trained. We collected photos from internal data, structuring the collection process by countries, cities and landmarks. This structure is as follows: we have divided the world into several regions, such as Europe (Russia included), America, Middle East, Africa, Far East, Australia and Oceania. In each region were selected cities in which there are a lot of significant landmarks. The significance of a landmark is determined by a combination of factors, such as popularity and availability of a sufficient number of photos. When collecting data, we focused mainly on those areas that are most popular among users. In addition, we filtered some landmarks out: not all parks and beaches are included in the database, but only the most popular and largest. This filtering was done because natural objects are difficult to distinguish. After selecting the cities of interest, in each of them significant landmarks were selected and photos for these landmarks were collected. Examples of images from our database is shown in Figure~\ref{fig:data_parts}. We collected not only images, but also meta data, which is used during the inference stage.

\subsubsection{Database cleaning} 
After collecting the database, we had to clean it before training a CNN. The raw database contained a lot of redundant images. For example, for Notre-Dame de Paris we need only photos of its facade, but not the interior. In addition, the raw data contained many erroneous images.

To clean the database, we picked up $5$ images per landmark, with a high probability of containing this landmark, and checked them manually. This is the only manual step of the algorithm. These resulting $3-5$ images called {\it landmark references}. We cleaned the database gradually, by parts used in the curriculum learning procedure from~\ref{seq:curr_learnig}: the landmark CNN trained in the previous step is used to clean each new part of the database. For each landmark we calculate a centroid from its references. This centroid is compared to each image of current landmark data using dot product similarity, and if the resulting similarity is less than the threshold $\gamma$, the image is rejected. More formally, the database cleaning procedure is written in the Algorithm 3.

\begin{algorithm}
\caption{Database cleaning algorithm}
\begin{algorithmic}[1]
\Statex \textbf{Input:} Landmark Images set; $M$ --- number of images in the set; Landmark CNN; Landmark References.
\Statex \textbf{Output:} Clean set of images.
\State \textit{references-embeddings} = CNN(References) --- pass landmark references through the landmark CNN and get embeddings
\State \textit{references-centroid} = average(\textit{references-embeddings})
\State \textbf{for} ($i \in [1, \ldots, M]$ ) \textbf{do}
\State \quad \textit{embedding} = CNN(Images[$i$])
\State \quad \textbf{if} (dotProduct(\textit{references-centroid}, \textit{embedding})) < $\gamma$) \textbf{then}
\State \quad \quad \textbf{yield} Reject
\State \quad \textbf{yield} Accept
\end{algorithmic}
\end{algorithm}

As a result, after cleaning our large scale database includes $11381$ landmarks from $503$ cities and $70$ countries, the total number of landmarks images in database is $2331784$. As non-landmarks we use about $900000$ images. The number of images per landmark is in range $20-1000$. The landmarks, the number of images for which less than $100$, we call ``rare'', the rest --- ``frequent''. For ``rare'' objects, due to the small amount of data, it is more difficult to train the model, therefore, we will separately consider the behavior on some tests of ``rare'' and ``frequent'' classes.

It's must be noted that for cleaning of 4 million images, manual checking required $56 905$ references images. Everything else has been done automatically.

\section{Experiments and analysis} \label{experiments}

In this section we present the results of experiments of our deployed landmark recognition system. We analyze model behavior in offline test, which approximates the real-world distribution of photos with landmarks, and in emulated online test. In addition, this section presents the results of deploying the system at Cloud Mail.ru and the performance of our model on the standard image retrieval test Revisited Paris~\cite{radenovic2018revisiting}.

\subsection{Quantitative experiments}

\subsubsection{Offline test}
In order to measure the quality of the model, we collected and manually labeled the offline testset. According to our calculations, on average, photos containing landmarks make up $1-3 \%$ of the total number of photos in the user's cloud. In our offline test we emulate this distribution, so this test contains $290$ images with landmarks and $11000$ images without them. It is important to note that a test photo may contain several landmarks and we decide that the model answered correctly, if its top 1 answer is among those landmarks. During the offline test geo information and landmark references weren't used. 

The results of this test are presented in Table~\ref{tab:offline_no_geo}. To measure the results of experiments, we used two metrics: {\it sensitivity} --- accuracy of a model on images with landmarks (also called Recall) and {\it specificity} --- accuracy of a model on images without landmarks. Our main model with non-landmark threshold $\eta=250$ (picked up on validation data) shows the best result, similar to state-of-the-art DELF~\cite{radenovic2018revisiting}. We experimented with several types of DELF and included best results in terms of sensitivity and specificity in Table~\ref{tab:offline_no_geo}. 

The table also contains the results of our model trained only with Softamx loss, with Softmax and Center loss, without curriculum learning and with using one centroid per class for inference. Thus, Table~\ref{tab:offline_no_geo} reflects improvements of our approach with the addition of new elements in it.

\begin{table}
  \caption{Offline test for various hyperparamters}
  \label{tab:offline_hyperparams}
  \begin{tabular}{lcc}
    \toprule
    \multirow{2}{*}{Hyperparamter}&Sensitivity&Specificity\\
    & \% & \%\\
    \midrule
    Non-landm. prop. = 0.2, $\eta = 350$ & 76 & 98.8\\
    {\bf Non-landm. prop. = 0.4, $\eta = 250$} & {\bf 80} & {\bf 99}\\
    Non-landm. prop. = 0.7, $\eta = 350$ & 74 & 99\\
    \hline
    256 embedding size, $\eta = 200$ & 75 & 99\\
    {\bf 512 embedding size, $\eta = 250$} & {\bf 80} & {\bf 99}\\
    1024 embedding size, $\eta = 200$ & 60 & 98.1\\
    \hline
    Single linkage, $\eta = 250$ & 76 & 98.9\\
    {\bf Complete linkage, $\eta = 250$} & {\bf 80} & {\bf 99}\\
    Average linkage, $\eta = 250$ & 78 & 98.8 \\
    \hline
    Clustering thresh. = 3, $\eta = 250$ & 74 & 99\\
    Clustering thresh. = 7, $\eta = 250$ & 76 & 98.7\\
    {\bf Clustering thresh. = 10, $\eta = 250$} & {\bf 80} & {\bf 99}\\
    Clustering thresh. = 15, $\eta = 250$ & 77 & 98.9\\
    \hline
    Min elem. per clust. = 10, $\eta = 250$ & 65 & 99 \\
    Min elem. per clust. = 25, $\eta = 250$ & 74 & 99 \\
    {\bf Min elem. per clust. = 50, $\eta = 250$} & {\bf 80} & {\bf 99}\\
    Min elem. per clust. = 100, $\eta = 250$ & 78 & 98.9\\
  \bottomrule
\end{tabular}
\end{table}

\begin{table}
  \caption{Offline test for kNN performance}
  \label{tab:offline_knn}
  \begin{tabular}{lcc}
    \toprule
    \multirow{2}{*}{kNN type}&Sensitivity&Specificity\\
    & \% & \%\\
    \midrule
    kNN embeddings, $k=1$, ($\eta=300$) & 76 & 99\\
    kNN embeddings, $k=5$, ($\eta=300$) & 77.8 & 98.8\\
    kNN embeddings, $k=10$, ($\eta=300$) & 78.8 & 99\\
    kNN embeddings, $k=50$, ($\eta=300$) & 77.8 & 99\\
    1 centroid & 62 & 98.8\\
    {\bf Several centroids, $k=1$, ($\eta=250$)} & {\bf 80} & {\bf99}\\
    Several centroids, $k=3$, ($\eta=250$) & 74 & 98.7\\
    Several centroids, $k=5$, ($\eta=250$) & 64 & 99\\
  \bottomrule
\end{tabular}
\end{table}

\begin{table}
  \caption{Offline test without geo info}
  \label{tab:offline_no_geo}
  \begin{tabular}{lcc}
    \toprule
    \multirow{2}{*}{Model}&Sensitivity&Specificity\\
    & \% & \%\\
    \midrule
    Softmax loss & 17 & 99\\
    Softmax + Center loss & 31 & 99 \\
    Ours (without curriculum learning) & 55 & 98\\
    Ours (1 centroid per class) & 62 & 98.8 \\
    {\bf Ours ($\eta=250$)} & {\bf 80} & {\bf99 }\\
    DELF~\cite{noh2017largescale} (5 candidates) & 77.8 & 99\\
    {\bf DELF~\cite{noh2017largescale} (20 candidates)} & {\bf 80.1} & {\bf 99}\\
  \bottomrule
\end{tabular}
\end{table}

\begin{figure*}
  \centering
  \includegraphics[width=\linewidth]{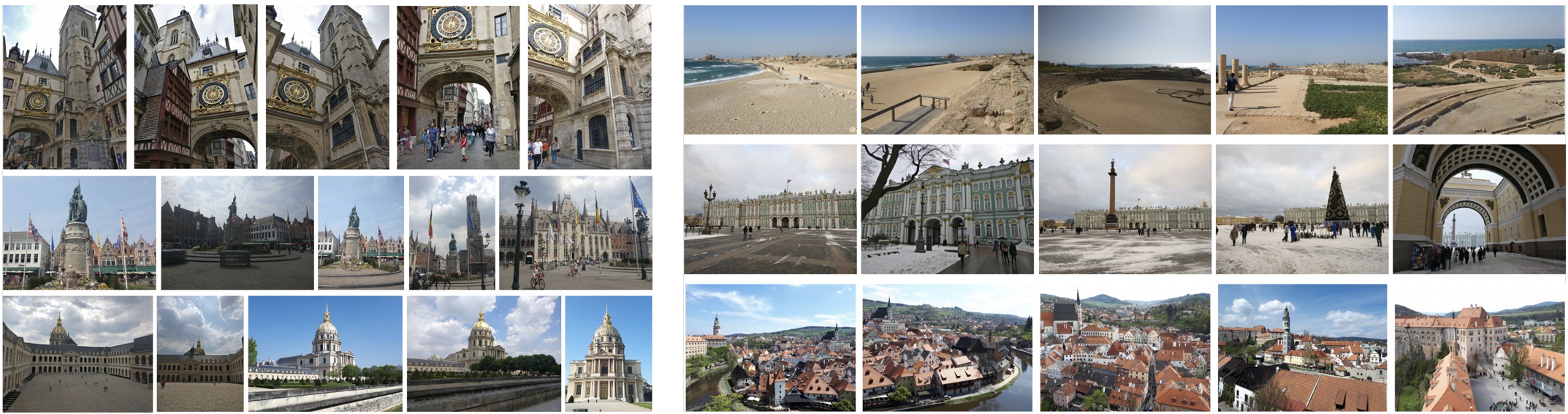}
  \caption{({\it Left}:) Top 5 images, closest to corresponding centroid. ``Rare'' classes. (Top row): Gros-Horloge in Rouen, (middle row): Jan Breydel and Pieter de Coninck Monument in Bruges, (bottom row): Esplanade des Invalides in Paris.\\
  ({\it Right}:) Top 5 images, closest to corresponding centroid. ``Frequent'' classes. (Top row): Caesarea National Park, (middle row): State Hermitage Museum and Winter Palace in St. Petersburg, (bottom row): Historic Center of Cesky Krumlov.}
  \Description{Top 5 ``rare'' and ``frequent''  results.}
  \label{fig:cloud_rare_freq}
\end{figure*} 

In Table~\ref{tab:offline_no_geo_rare_freq}, we examine the behavior of our model separately on ``rare'' and ``frequent'' landmarks. Column ``Part from total number'' shows what percentage of landmark examples in the offline test has the corresponding type of landmarks. It's very important to understand how model works on ``rare'' landmarks due to the small amount of data for them.

\begin{table}
  \caption{``Rare'' / ``frequent'' offline test without geo info}
  \label{tab:offline_no_geo_rare_freq}
  \begin{tabular}{ccc}
    \toprule
    \multirow{2}{*}{Landmark type}&Sensitivity&Part from total number\\
    & \% & \%\\
    \midrule
    ``Rare'' & 46 & 13 \\
    ``Frequent'' & 85.3 & 87\\
  \bottomrule
\end{tabular}
\end{table}

Analysis of the behavior of our model in different categories of landmarks in the offline test is presented in Table~\ref{tab:offline_no_geo_categories}. These results show that the model can successfully work with various categories of landmarks. 

\begin{table}
  \caption{Landmarks categories offline test without geo info}
  \label{tab:offline_no_geo_categories}
  \begin{tabular}{lcc}
    \toprule
    \multirow{2}{*}{Category}&Sensitivity&Part from\\
    & \% & total number \%\\
    \midrule
    Museums & 83 & 10\\
    Historic sites & 84 & 10\\
    Churches and Cathedrals & 83 & 11\\
    Architectural buildings & 74 & 7 \\
    Gardens and Parks & 84 & 5\\
    Monuments and Statue & 90 & 4 \\
  \bottomrule
\end{tabular}
\end{table}

When we launched our model on the offline test with geo information, we got predictably better results: $92 \%$ of sensitivity and $99.5 \%$ of specificity. It's must be noted that our test in the non-landmark part contains photos, taken close to landmarks in order to pass geographic filtering step.

\subsubsection{Emulated online test}
In order to test behavior of our model on real-world data, we run it on the set of $581545$ images. As a result, about $3 \%$ of images containing landmarks were found, which corresponds to our preliminary estimates of the distribution of landmarks. Figure~\ref{fig:cloud_rare_freq} shows top 5 closest images to corresponding centroids. Left part of Figure~\ref{fig:cloud_rare_freq} contains results for ``rare'' landmarks classes, right part --- for ``frequent'' classes. All shown images are donated by colleagues.

\subsubsection{Revisited Paris dataset}

We measured quality of our landmark recognition approach on Revisited Paris dataset ($\mathcal{R}$Par)~\cite{radenovic2018revisiting}. This dataset with Revisited Oxford ($\mathcal{R}$Oxf)~\cite{radenovic2018revisiting} are standard benchmarks for comparison of image retrieval algorithms. Despite the fact that the retrieval and recognition are similar tasks, the recognition task is broader. In the formulation of the retrieval problem, it is necessary to find in the database images that are close to the query image. In the recognition we have to determine the landmark, which is contained in the query image. Images of the same landmark can have different shooting angles or taken inside/outside the building.

For this reason, we measured the quality of our model  in the standard and adapted to our task settings. Not all classes from queries are presented in our landmark dataset. For Revisited Oxford there are only 4 landmarks, so we did not consider this test. For $\mathcal{R}$Par in only our classes setting remains 59 queries out of 70. In $\mathcal{R}$Par there are images that clearly contain the correct landmark, but taken from different shooting angles or from inside the building. We transferred this images from the category ``wrong'' to the ``junk'' category. This category do not influence on the final score. Such ``junk revision'' setting makes the test markup closer to the task that our model solves.

Results on $\mathcal{R}$Par with and without distractors in medium and hard modes are presented in Tables~\ref{tab:paris_medium} and \ref{tab:paris_hard}. We also added results of methods from~\cite{radenovic2018revisiting} and \cite{teichmann2018detect} in best settings to these tables.

\begin{table}
  \caption{Revisited Paris Medium}
  \label{tab:paris_medium}
  \begin{tabular}{l|cc|cc|}
  \multirow{2}{*}{Method} & \multicolumn{2}{c}{$\mathcal{R}$Par} & \multicolumn{2}{c}{$\mathcal{R}$Par + $\mathcal{R}$1M}\\
  & mAP & mP@10 & mAP & mP@10\\
  \hline
   AlexNet-GeM \cite{radenovic2018fine} & 58.0 & 91.6 & 29.9 & 84.6\\
   VGG16-GeM \cite{radenovic2018fine} & 69.3 & 97.9 & 45.4 & 94.1\\
   ResNet101-GeM \cite{radenovic2018fine} & 77.2 & 98.1 & 52.3 & 95.3\\
   ResNet101-R-MAC \cite{gordo2017end} & 78.9 & 96.9 & 54.8 & 93.9\\
   HesAff (best) \cite{tolias2016image} & 61.4 & 97.9 & 42.3 & 96.3\\
   DELF (best)\cite{noh2017largescale} & 76.9 & 99.3 & 57.3 & 98.3\\
   DELF---GLD (best)\cite{teichmann2018detect} & 80.2 & 99.1 & 59.7 & 99.0\\ 
   \hline
   Ours (standard setting) & 70.0 & 91.0 & 54.8 & 81.6\\
   Ours (our classes) & 75.5 & 93.6 & 63.2 & 88.9\\
   Ours (revised junk) & 80.1 & 95.3 & 60.3 & 84.1\\
   Ours (revised junk \\+ our classes) & 86.0 & 97.5 & 69.5 & 91.7\\
\end{tabular}
\end{table}

\begin{table}
  \caption{Revisited Paris Hard}
  \label{tab:paris_hard}
  \begin{tabular}{l|cc|cc|}
  \multirow{2}{*}{Method} & \multicolumn{2}{c}{$\mathcal{R}$Par} & \multicolumn{2}{c}{$\mathcal{R}$Par + $\mathcal{R}$1M}\\
  & mAP & mP@10 & mAP & mP@10\\
  \hline
   AlexNet-GeM \cite{radenovic2018fine} & 29.7 & 67.6 & 8.4 & 39.6\\
   VGG16-GeM \cite{radenovic2018fine} & 44.3 & 83.7 & 19.1 & 64.9\\
   ResNet101-GeM \cite{radenovic2018fine} & 56.3 & 89.1 & 24.7 & 73.3\\
   ResNet101-R-MAC \cite{gordo2017end} & 59.4 & 86.1 & 28.0 & 70.0\\
   HesAff (best) \cite{tolias2016image} & 35.0 & 81.7 & 16.8 & 65.3\\
   DELF \cite{noh2017largescale} & 55.4 & 93.4 & 26.4 & 75.7\\
   DELF---GLD (best) \cite{teichmann2018detect} & 58.6 & 91.0 & 29.4 & 83.9\\ 
   \hline
   Ours (standard setting) & 50.2 & 70.9 & 34.9 & 59.7\\
   Ours (our classes) & 54.5 & 75.9 & 40.2 & 68.1\\
   Ours (revised junk) & 68.2 & 89.0 & 43.3 & 73.1\\
   Ours (revised junk \\+ our classes) & 74.0 & 95.4 & 50.1 & 83.7\\
\end{tabular}
\end{table}
 


\subsection{Deployment results}

Our landmark recognition system has been deployed deployed and used to process user photos in the photo storage app Cloud Mail.ru. If there are any landmarks in the photo, they are tagged and categorized into corresponding albums, which are in turn grouped by cities and countries. The example of the mobile application interface is shown in Figure~\ref{fig:intro}. These albums help users to easily find and explore photos with landmarks.

To measure the impact of the landmark recognition system on user behavior, it was split-tested on $10 \%$ of users. The system processed several billion photos, of which were registered 31 million photos with landmarks on them. 

During the month of comparison of two users' groups with and without landmark recognition, the following changes occurred: purchase conversion increased by $10\%$, mobile app removal conversion decreased by $3 \%$, the number of views of albums increased by $12\%$.

\section{Comparison} \label{sec:comparison}

We compared the system, presented in this paper, to other landmark recognition methods. Recent most efficient approaches as well as ours are built on fine-tuned convolutional neural networks~\cite{gordo2017end}, \cite{radenovic2018fine}, \cite{noh2017largescale}. Based on the conclusions made in~\cite{radenovic2018revisiting}, we chose to compare it to DELF~\cite{noh2017largescale} as the state-of-the-art approach.

As a result of the experiments, we noticed that DELF depends very much on the cleanliness of the landmark database due to the use of deep local features through kNN search. Centroids also depend on how clean the database they were built on is; however, they are more robust and resistant to outliers. We experimented with incorporation of the attention-based method (like in DELF) into our trained landmark CNN. But we got pretty much the same results and came to the conclusion that the CNN has already learned and captured everything we try to with the attention mechanism.

\subsection{Offline test}
Comparison of our approach and DELF on our offline test is presented in Table~\ref{tab:offline_no_geo}. During this test DELF used the same database for retrieval. For our experiments we used official DELF implementation \footnote{https://github.com/tensorflow/models/tree/master/research/delf}. As can be seen from the table, both approaches show similar results.

During the kNN search stage, for each feature of a query image DELF identifies $60$ nearest features in the search index. Images with the highest number of matched features are considered candidates. 
We experimented with various DELF's parameters and got best results in terms of sensitivity and specificity using $20$ candidates.

\subsection{Timing}
We also compare the inference speed of our approach and DELF's. During inference stage DELF uses 7 image scales (from $0.25$ to $2$), which leads 7 forward passes through the CNN for one image; our method needs only one pass per image. Both approaches use kNN index search; we used Faiss for both. Since DELF searches the entire database (which in our case consists of more than 2 millions images), this step takes about $0.32$ second. Our method uses only $15250$ centroids and needs only $0.05$ second to yield final results. DELF further performs RANSAC method for geometric verification which is very time consuming: for $20$ candidates it takes $0.428$ second. We experimented with a smaller number of candidates, $5$ for example, and got an acceleration of up to $0.1$ second, however, led to lower accuracy (see Table~\ref{tab:offline_no_geo} for comparison). In total, our method needs $0.05$ second for inference (without the CNN pass), and DELF needs about $0.748$ --- 15 times slower. Due to the fact that our approach uses centroids, it needs to store only $15250$ elements, while DELF needs to store about 2 millions resulting in more than 133 times memory savings in our setup. 

All measurements were performed on 2x2660 Intel CPU, 128Gb Memory and SSD disks.

To sum up, the proposed approach yields same sensitivity / specificity results, performs just one forward pass on GPU (instead of 7), is faster 15x on CPU inference stage and by order of magnitude more memory-efficient.

\section{Conclusion} \label{sec:conclusion}

In this paper we proposed scalable approach for landmark recognition. Its implementation has been deployed to production at scale and used for recognition on user photos in Mail.ru cloud application. The basis of the method is the usage of embeddings of the deep convolutional neural network. This CNN is trained by curriculum learning technique with modified version of Center loss. 
To decide whether a landmark is present on a given image we compute distance between the image embedding vector and one or more centroids per class. For each landmark the centroids are calculated over clusters obtained from hierarchical clustering procedure. In addition, we presented a method for cleaning landmark data. Our approach shows similar results to the state-of-the-art-method, but it's much faster and suitable for deploying at scale.

\begin{acks}
The authors would like to thank the entire Vision Mail.ru team for their contributions to the project. We would like to thank Yan Romanikhin for fruitful discussions, Evgeniy Zholkovskiy for help with some experiments, Alan Basishvili, Taras Miskovec and Sergey Tarasenko for helpful advice.
\end{acks}

\bibliographystyle{ACM-Reference-Format}
\bibliography{biblio}

\end{document}